\documentclass{article}
\usepackage[sectionbib]{chapterbib}

\usepackage{graphicx}

\usepackage[preprint]{nips_2018}

\usepackage[utf8]{inputenc} 
\usepackage[T1]{fontenc}    
\usepackage{hyperref}       
\usepackage{url}            
\usepackage{booktabs}       
\usepackage{amsfonts}       
\usepackage{nicefrac}       
\usepackage{microtype}      
\usepackage{multirow}
\usepackage{color}
\usepackage{natbib}
\begin{document}

\title{A simple blind-denoising filter inspired by electrically coupled photoreceptors in the retina}

%

\author{
  Yang Yue \\
  Peking University\\
  \texttt{yueyang999@pku.edu.cn} \\
  \And
  Liuyuan He \\
  Peking University\\
  \texttt{liyhe@pku.edu.cn} \\
  \And
  Gan He \\
  Peking University\\
  \texttt{hegan@pku.edu.cn} \\
  \And
  Jian.K.Liu \\
  University of Leicester\\
  \texttt{jian.liu@leicester.ac.uk} \\
  \And
  Kai Du$^{*}$ \\
  Department of Neuroscience \\
  Karolinska Institutet\\
  \texttt{Kai.Du@ki.se} \\
  \And
  Yonghong Tian$^{*}$ \\
  Peking University\\
  \texttt{yhtian@pku.edu.cn} \\
  \And
  Tiejun Huang \\
  Peking University\\
  \texttt{tjhuang@pku.edu.cn} \\
}
\maketitle
\begin{abstract}
Photoreceptors in the retina are coupled by electrical synapses called “gap junctions”. It has long been established that gap junctions increase the signal-to-noise ratio of photoreceptors. Inspired by electrically coupled photoreceptors, we introduced a simple filter, the PR-filter, with only one variable. On BSD68 dataset, PR-filter showed outstanding performance in SSIM during blind denoising tasks. It also significantly improved the performance of state-of-the-art convolutional neural
network blind denosing on non-Gaussian noise. The performance of keeping more details might be attributed to small receptive field of the photoreceptors.
\end{abstract}

\section{Introduction}
Photoreceptors are the first stage of our early visual system, which transfer lights signals impinged on the retina into electrical signals\cite{masland2012neuronal,demb2015functional}. The photoreceptors produce large amount of noises during phototransduction process\cite{rieke1998single}. Through millions of years of evolution, our retina has developed a simple strategy for denoising—making photoreceptors electrically coupled via synapses called “gap junctions”\cite{devries2002electrical,evans2002gap,asteriti2014mouse,asteriti2015cambrian,asteriti2015slow}. It was proposed that photoreceptor electrical couplings are able to increase signal-to-noise ratio and important for noise filtering\cite{devries2002electrical,evans2002gap,li2012gap}. However, the denoising mechanism for large-scale, electrically coupled photoreceptor network remains unexplored. Importantly, the strategy of making our biologic “photo-sensors” electrically coupled is entirely different from electronic photo-sensors in cameras or other devices, which must be strictly isolated from each other. 

Here, we explored the denoising effects in the photoreceptor network with gap junctions. The aims of this study are multiple: (1) to gain insights into denoising mechanisms of electrical coupled photoreceptors in the retina; (2) to apply the biology-derived principle into machine denoising. 

The structure of this paper is organized as follows: First, we applied the “spike-triggered average (STA)” method\cite{chichilnisky2001simple,sharpee2013computational,touryan2005spatial} to estimate the receptive field of individual photoreceptors in a biologically detailed photoreceptors network. The STA analysis indicate single photoreceptor has a small receptive field (Fig.2C), suggesting single photoreceptor summarize information very locally. Next, we constructed a photoreceptor-filter (PR-filter) – a grid network with electrically coupled PRs. We extensively tested the PR-filter on various types of noise. The SSIM performance of the PR-filter exceeds any other classic spatial filters. At last, we combined the PR-filter with deep convolutional neural network (CNN). Our results indicates that the PR-filter can significantly improve the performance of the state-of-the-art CNN denoising on blind non-Gaussian noise\cite{zhang2017beyond}, not only increasing the signal-to-noise ratio but also keeping more details in vision. 

Over all, our work is the first to apply the gap junctions in the machine denoising tasks and the first to show the gap junctions can be directly integrated in the modern AI architecture, which is significant for understanding evolution emergence of eletrical couplings between photoreceptors and may inspire blind denoising in machine learning.

\section{Background}

\subsection{Photoreceptor noise}
Photoreceptors noises mainly originate from two parts\cite{rieke2000origin}: (1) dark noise produced by intracellular activities, (2) ambient illuminations. In low-light intensity conditions, the noise can be several log units higher than signals\cite{rieke1998single,barlow1993molecular}.  

\subsection{Gap junctions in the retina}
Gap junctions are electrical synapses which allow cells to directly communicate with electrical signals\cite{evans2002gap}. Retina has a layer-wised structure, while photoreceptors (rods and cones) are located in the first layer (Fig.1). Gap junctions are abundant in each layers of the retinal (Fig.1), suggesting the gap junction is fundamental to retina signal processing\cite{bloomfield2009diverse}. 

\begin{figure}[htbp]
  \centering
  \includegraphics[width=6cm]{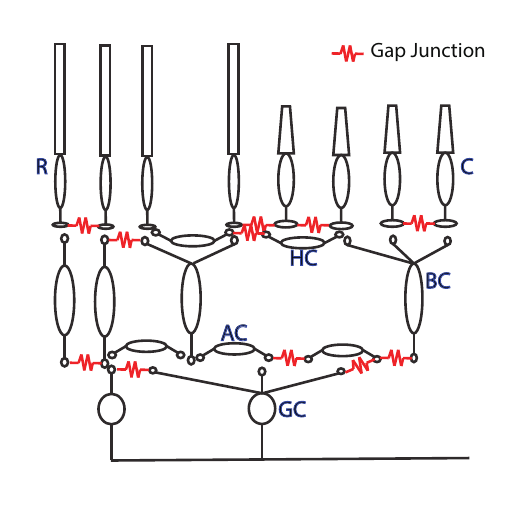}
  \caption{Gap junctions in the retina circuity. Information of light stimulus flow through three layers of neurons: cone (C) and rod (R) as photoreceptors, cone and rod bipolar cells (BC), ganglion cells (GC).}
  \label{fig1}
\end{figure}

\subsection{Receptive field and STA analysis}
In neuroscience, “receptive field” refers to a specific region of stimuli (in space) that can affect a neuron’s response\cite{theunissen2001estimating}. The STA analysis is one of the most popular methods which has been widely used in studying receptive field in visual and auditory system\cite{chichilnisky2001simple,simoncelli2004characterization}. The STA was initially designed for spiking neurons, but it can also apply to non-spiking neurons, such as BC\cite{liu2017inference}. For estimating non-linear receptive field, one should refer to other methods such as spike-triggered covariance (STC)\cite{paninski2003convergence} and Bayesian method\cite{park2011bayesian}. 

To compute the STA, the whole stimulus time is divided into $k$ bins equally. Let $x_{i}$ denotes the stimulus vector preceding ith bin and here the mean of stimuli are assumed to be zero. And $y_{i}$ denotes the spike number in $i_{th}$ bin. The STA is given by
\begin{equation}
STA=\frac{1}{n_{s}}\sum_{i=1}^{n_{s}} y_{i} x_{i}\qquad
\end{equation}
where $n_{s}$ is the total number of spikes. The STA method requires stimulus to be strictly “spherically symmetric”, such as  white noise\cite{sharpee2004analyzing,bussgang1952crosscorrelation}. If the stimulus distribution is not spherically symmetric, ZCA whitening can be applied before computing STA. 

\section{Experiments} 

\subsection{STA analysis of electrically coupled photoreceptors}

\begin{figure}[htbp]
  \centering
  \includegraphics[width=10cm]{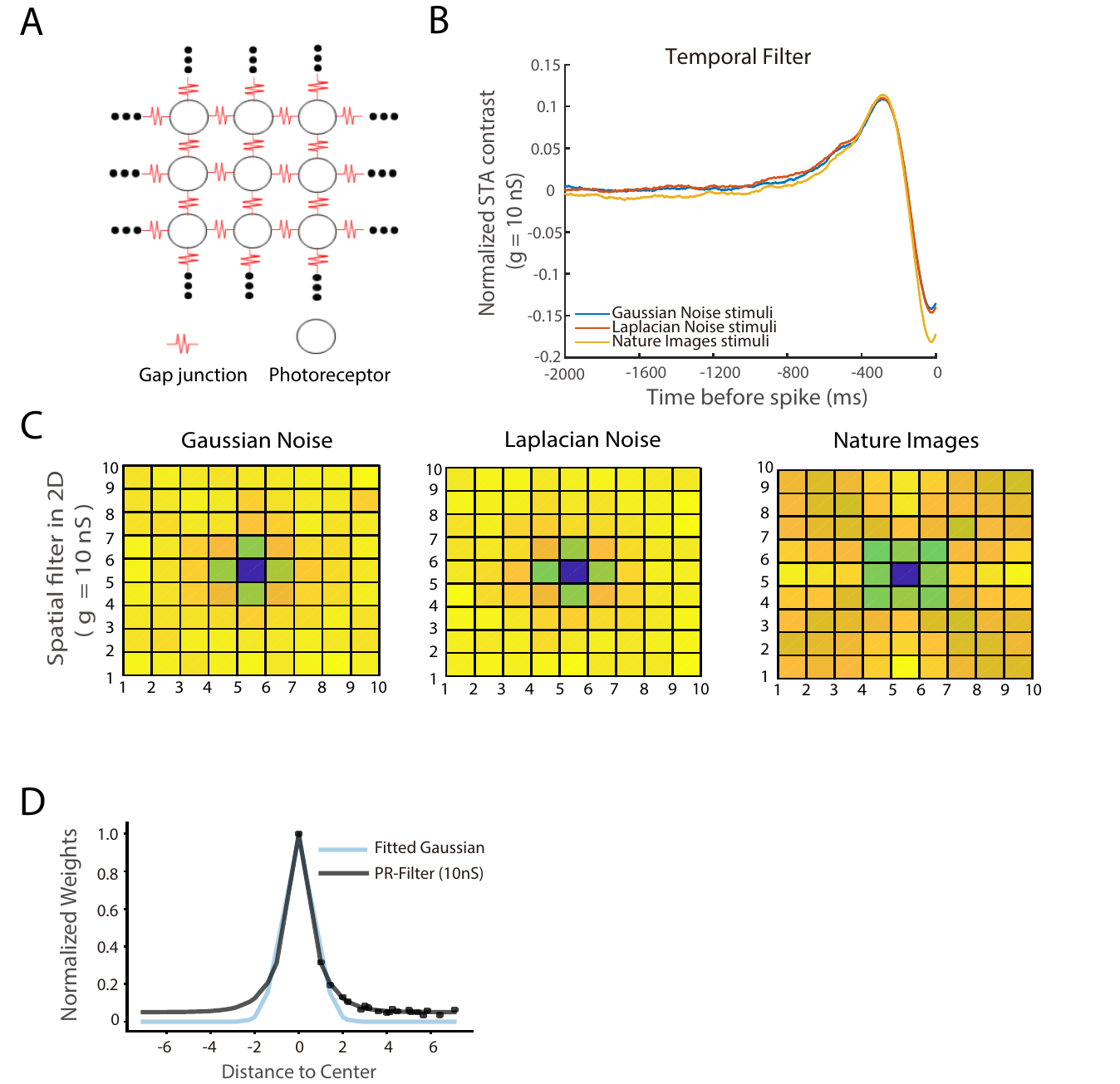}
  \caption{ (A) Schematic diagram for 10x10 Photoreceptor network. Photoreceptors are organized as grid and gap junctions connect them as edges. (B) Normalized weights in temporal filters. The negative peaks appear at $\sim$24 ms prior to the spikelet, indicating stimulus perturbations at this time point, on average, make maximal negative correlation to the spikelet. (C) STA spatial filters(receptive fields) under different stimuli for the center cell in network. White Gaussian Noise is labeled as ‘Gaussian Noise’. Laplacian noise and nature images are whitened before fed into model. When corresponding line in (B) reaches its minimum value, we computed STAs for all photoreceptors in the network and “froze” them at $\sim$24 ms prior to the spikelet, which reveal the spatial receptive field for the centered neuron in the network. (D) PR filter compares with Gaussian fliter.}
  \label{fig2}
\end{figure}

To gain insights into the functional role of gap junctions at the network level, we investigated the spatiotemporal receptive field of photoreceptors in a network, wherein gap junction couplings are the only connections among cells. We constructed a biologically detailed model for the photoreceptor, including phototransduction cascade and ion channels (Appendix Fig.A1.1, Table.A1.1, Table.A1.2, and Table.A1.3). The model is described as follows:
\begin{equation}
C_{m}\frac{\mathrm{d}v_{m}}{\mathrm{d}t}=\sum g_{ion}(v_{m}-e_{ion})+g_{leak}(v_{m}-e_{leak})+I_{gap}+I_{photo}\qquad 
\end{equation}
\begin{equation}
I_{gap}=g_{gap}(v_{m}-v_{gap})
\end{equation}
where $v_{m}$ is the membrane voltage of photoreceptor, $C_{m}$ is the membrane voltage capacitance, $v_{gap}$ is the voltage of gap junction, $I_{gap}$ and $I_{photo}$ are currents from gap junction and phototransduction respectively, $e_{ion}$ and $e_{leak}$ are the voltage potentials of ion channel and leak, $g_{leak}$ is leak conductance, and $g_{ion}$ is ion channel conductance in Hodgkin-Huxley form\cite{hodgkin1952quantitative}.
The photoreceptor model can generate photocurrents in close match to available electrophysiology data (Appendix Fig.A1.2)\cite{baylor1986electrical} . The topology of the network is grid-like ($10\times10$ in size) and consisted of 100 photoreceptors (Fig.2A). In such a network, a neuron can only form connections to neighboring neurons and there is no long-range connections\cite{wells2016variation}. All simulations were performed in the NEURON simulator\cite{hines1997neuron}. 

To first investigate receptive field of photoreceptors under different stimuli, we feed the network with three types of inputs: Gaussian noise, Laplacian noise and natural images. Fig.2B shows the temporal STAs computed at the center neuron of the network. For all tested stimuli, our results show “off-center” receptive field of the center neuron, suggesting gap junctions provide inhibition-like effects at this moment (Fig.2C). Interestingly, the shapes of the spatial receptive field are similar to a narrow Gaussian filter($\sigma$=0.7358)(Fig.2D), but with a heavier "tail", suggesting that the receptive field of single photoreceptor is largely weighted on the local connections, but sum up more information from larger area compared to a narrow Gaussian filter.

\subsection{PR-filter on blind-denoising}

\begin{figure}[htbp]
  \centering
  \includegraphics[width=12cm]{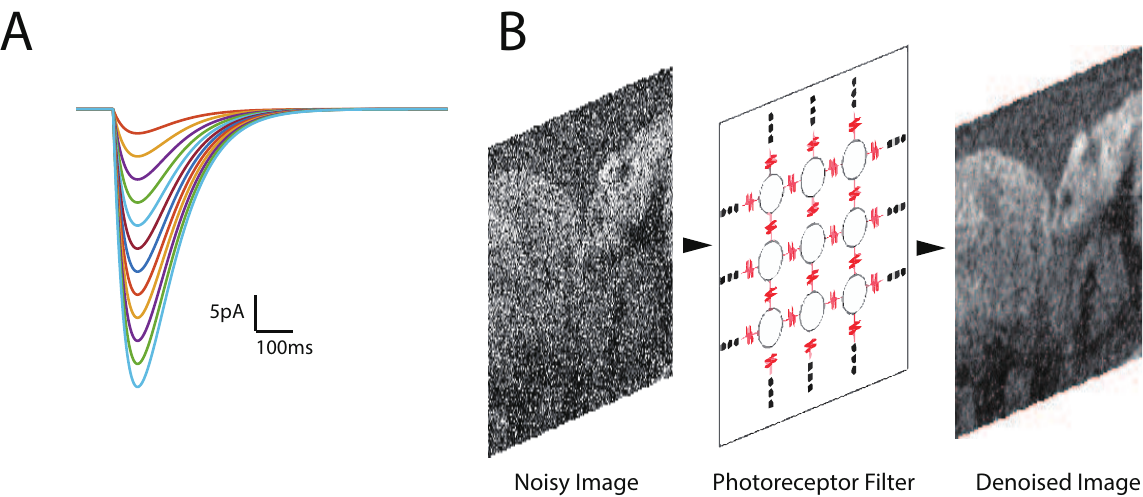}
  \caption{(A) Simplified photocurrent curves in the PR-filter. (B) Value of pixels in input image is converted into stimuli for network model and the peak values of the voltage responses is normalized as the output image.}
  \label{fig3}
\end{figure}

\begin{table}[htbp]
\scriptsize
  \caption{All PSNR comparison of single filter}
  \label{table1}
  \centering
  \begin{tabular}{cccccccccc}
    \toprule
    Noise Type	& noisy 	&Adaptive Median	& Average	& Gaussian	& Max	& Mean	& Median	& Min	& \emph{PR} \\
    \midrule
\multirow{4}*{Gaussian} &9.3931	&12.3257	&16.4323	&16.7648	&5.6420	&16.8155	&14.7134	&7.4840	&\color{red}\emph{19.4910}\\
&12.0365	&14.9079	&19.1847	&19.6653	&7.6740	&19.6792	&18.1044	&8.9699	&\color{red}\emph{20.9816}\\
&14.9113	&17.6838	&21.5659	&\color{red}22.2076	&10.2093	&22.1672	&20.9178	&10.9068	&\emph{21.7804}\\
&18.1941	&20.7064	&23.6390	&\color{red}24.6108	&12.9497	&24.3491	&23.4640	&13.2402	&\emph{23.3808}\\
\hline
\multirow{4}*{I.D.G.} &9.3962	&12.3250	&16.4365	&16.7691	&5.6442	&16.8200	&14.7149	&7.4880	&\color{red}\emph{19.5363}\\
&11.6292	&14.5065	&18.8063	&19.2645	&7.3250	&19.2853	&17.6505	&8.7198	&\color{red}\emph{20.6644}\\
&14.9198	&17.6923	&21.5705	&\color{red}22.2112	&10.2112	&22.1714	&20.9270	&10.9191	&\emph{21.7553}\\
&17.7218	&20.2876	&23.3770	&\color{red}24.2926	&12.5780	&24.0730	&23.1347	&12.9066	&\emph{23.1438}\\
\hline
\multirow{4}*{Laplacian} &9.5985	&13.2473	&16.6411	&16.9802	&5.5888	&17.0283	&15.8758	&7.4579	&\color{red}\emph{19.4907}\\
&12.0132	&15.7000	&19.1456	&19.6166	&7.2323	&19.6343	&19.3830	&8.7341	&\color{red}\emph{20.8815}\\
&14.4932	&18.2277	&21.2598	&21.8724	&9.2729	&21.8447	&\color{red}21.9666	&10.3296	&\emph{21.7843}\\
&17.8840	&21.4520	&23.4909	&24.4264	&12.1308	&24.1948	&\color{red}24.4911	&12.6902	&\emph{23.2115}\\
\hline
\multirow{4}*{Salt\&Pepper} &9.6453	&\color{red}28.3222	&16.5263	&16.8384	&5.2340	&16.8919	&19.8959	&7.3248	&\emph{18.9588}\\
&12.0806	&\color{red}30.7300	&19.0877	&19.5262	&6.4557	&19.5544	&25.8218	&8.5177	&\emph{20.2686}\\
&15.0880	&\color{red}32.2882	&21.6591	&22.2638	&8.4892	&22.2465	&28.2621	&10.4893	&\emph{21.2754}\\
&18.0933	&\color{red}32.8780	&23.6104	&24.5444	&10.7724	&24.3107	&28.8849	&12.6374	&\emph{22.8880}\\
\hline
\multirow{4}*{Uniform} &8.8903	&11.1816	&15.8579	&16.1635	&5.6221	&16.2197	&13.1821	&7.4633	&\color{red}\emph{19.1190}\\
&11.7585	&13.7830	&18.9400	&19.4126	&8.1607	&19.4283	&16.5059	&9.1614	&\color{red}\emph{20.8624}\\
&14.3920	&16.2600	&21.1588	&\color{red}21.7747	&10.5340	&21.7410	&19.0915	&11.0152	&\emph{21.6886}\\
&17.6805	&19.3204	&23.3418	&\color{red}24.2517	&13.2001	&24.0344	&21.9098	&13.3752	&\emph{23.0765}\\
\hline
\multirow{4}*{Blind} &9.5785	&13.9324	&16.3319	&16.6370	&5.4844	&16.6862	&16.1899	&7.4002	&\color{red}\emph{19.4510}\\
&12.4169	&17.3056	&19.3707	&19.8453	&7.2606	&19.8553	&20.2933	&8.8709	&\color{red}\emph{21.0081}\\
&15.2791	&20.3255	&21.7421	&22.3774	&9.4325	&22.3364	&\color{red}23.0744	&10.8026	&\emph{21.7454}\\
\hline
\multirow{4}*{Blind NG} &9.6814	&14.2938	&16.7078	&17.0474	&5.5096	&17.0955	&16.7126	&7.4115	&\color{red}\emph{19.5333}\\
&14.1590	&19.0992	&20.9855	&21.5724	&8.5531	&21.5534	&\color{red}22.5579	&9.9460	&\emph{21.4669}\\
&17.7802	&\color{red}28.5549	&23.4309	&24.3325	&10.6422	&24.1224	&27.7345	&12.4137	&\emph{21.5955}\\
    \bottomrule
  \end{tabular}
\end{table}

\begin{table}[htbp]
\scriptsize
  \caption{All SSIM comparison of single filter}
  \label{table2}
  \centering
  \begin{tabular}{cccccccccc}
    \toprule
    Noise Type	& noisy 	& Adaptive Median	& Average	& Gaussian	& Max	& Mean	& Median	& Min	& \emph{PR} \\
    \midrule
    \multirow{2}*{Gaussian}&0.0660	&0.0883	&0.1731	&0.2083	&0.1598	&0.2053	&0.1479	&0.0219	&\color{red}\emph{0.3969}\\
&0.1204	&0.1561	&0.2668	&0.3176	&0.1820	&0.3115	&0.2458	&0.0623	&\color{red}\emph{0.5023}\\
&0.1950	&0.2437	&0.3703	&0.4360	&0.2388	&0.4249	&0.3544	&0.1308	&\color{red}\emph{0.5695}\\
&0.3005	&0.3608	&0.4860	&0.5640	&0.3256	&0.5462	&0.4789	&0.2329	&\color{red}\emph{0.6299}\\
\hline
\multirow{2}*{I.D.G.}&0.0660	&0.0882	&0.1730	&0.2083	&0.1597	&0.2053	&0.1477	&0.0220	&\color{red}\emph{0.3976}\\
&0.1113	&0.1449	&0.2526	&0.3011	&0.1763	&0.2956	&0.2309	&0.0546	&\color{red}\emph{0.4884}\\
&0.1954	&0.2443	&0.3708	&0.4365	&0.2391	&0.4255	&0.3547	&0.1314	&\color{red}\emph{0.5694}\\
&0.2841	&0.3432	&0.4698	&0.5466	&0.3126	&0.5298	&0.4614	&0.2172	&\color{red}\emph{0.6215}\\
\hline
\multirow{2}*{Laplacian}&0.0702	&0.1150	&0.1795	&0.2158	&0.1447	&0.2127	&0.1816	&0.0219	&\color{red}\emph{0.4040}\\
&0.1217	&0.1869	&0.2653	&0.3162	&0.1538	&0.3098	&0.2976	&0.0537	&\color{red}\emph{0.4993}\\
&0.1870	&0.2733	&0.3566	&0.4206	&0.1882	&0.4099	&0.4112	&0.1012	&\color{red}\emph{0.5614}\\
&0.2955	&0.4038	&0.4778	&0.5554	&0.2601	&0.5378	&0.5482	&0.1864	&\color{red}\emph{0.6248}\\
\hline
\multirow{2}*{Salt\&Pepper}&0.0737	&\color{red}0.8839	&0.1805	&0.2167	&0.0775	&0.2135	&0.5545	&0.0284	&\emph{0.3980}\\
&0.1364	&\color{red}0.9237	&0.2720	&0.3241	&0.0800	&0.3161	&0.7846	&0.0596	&\emph{0.4936}\\
&0.2518	&\color{red}0.9374	&0.3930	&0.4624	&0.1196	&0.4462	&0.8313	&0.1233	&\emph{0.5638}\\
&0.4180	&\color{red}0.9397	&0.5148	&0.5972	&0.2093	&0.5711	&0.8409	&0.2291	&\emph{0.6225}\\
\hline
\multirow{2}*{Uniform}&0.0566	&0.0568	&0.1558	&0.1879	&0.1841	&0.1853	&0.1122	&0.0220	&\color{red}\emph{0.3725}\\
&0.1127	&0.1165	&0.2575	&0.3067	&0.2472	&0.3013	&0.1938	&0.0826	&\color{red}\emph{0.4958}\\
&0.1778	&0.1859	&0.3507	&0.4138	&0.3254	&0.4039	&0.2776	&0.1719	&\color{red}\emph{0.5601}\\
&0.2796	&0.2932	&0.4675	&0.5440	&0.4255	&0.5275	&0.3944	&0.3043	&\color{red}\emph{0.6215}\\
\hline
\multirow{2}*{Blind}&0.0681	&0.1377	&0.1737	&0.2088	&0.1293	&0.2058	&0.1870	&0.0263	&\color{red}\emph{0.3900}\\
&0.1341	&0.2387	&0.2790	&0.3324	&0.1422	&0.3247	&0.3300	&0.0633	&\color{red}\emph{0.4640}\\
&0.2215	&0.3513	&0.3878	&0.4563	&0.1878	&0.4425	&0.4624	&0.1306	&\color{red}\emph{0.5621}\\
\hline
\multirow{2}*{Blind NG}&0.0722	&0.1461	&0.1826	&0.2193	&0.1321	&0.2161	&0.2087	&0.0221	&\color{red}\emph{0.4061}\\
&0.1838	&0.3089	&0.3460	&0.4090	&0.1620	&0.3980	&0.4400	&0.0919	&\color{red}\emph{0.5300}\\
&0.3715	&0.7518	&0.4989	&0.5797	&0.2091	&0.5558	&\color{red}0.7612	&0.2065	&\emph{0.6192}\\
    \bottomrule
  \end{tabular}
\end{table}

\footnotetext[1]{Red mark stands for best performance of this line.}
\footnotetext[2]{Intensity denpendent Gaussian noise.}
\footnotetext[3]{Blind noise mixed by five regular noise types,which are white Gaussian,
intensity denpendent Gaussian(I.D.G.), Laplacian, Salt\&Pepper, and uniform.}
\footnotetext[4]{Blind noise mentioned above, but excluded two kinds of Gaussian noise.}
The STA analysis indicate the electrically coupled photoreceptor network can form self-organized receptive field, similar to a narrow Gaussian filter. We wonder if the photoreceptor network can be directly applied for image denoising. However, one problem in applying the biologically detailed network on high resolution images is the computation cost, due to simulated phototransduction cascade (Appendix Fig.A1). To this end, we simplified the network in the following procedure: (1) We removed all ion channels and only kept passive leak channels. The model was reduced to: 
\begin{equation}
C_{m}\frac{\mathrm{d}v_{m}}{\mathrm{d}t}=g_{leak}(v_{m}-e_{leak})+I_{gap}+I^{*}_{photo}
\end{equation}
(2) We used simplified photocurrents to replace the phototransduction cascade model. The photocurrents were reduced to double-exponential function: 
\begin{equation}
I^{*}_{photo}=I_{dark}-g_{max}\frac{\tau_{2}(e^{-\frac{t}{\tau_{1}}}-e^{-\frac{t}{\tau_{2}})}}{(\tau_{1}-\tau_{2}){\frac{\tau_{2}}{\tau_{1}}}^{\frac{\tau_{1}}{\tau_{1}-\tau_{2}}}}
\end{equation}
where $I_{dark}=40pA$, $tau_{1}=64ms$, $tau_{2}=68ms$, and $g_{max}$ controls the amplitude of $I^{*}_{photo}$.
We found the choice of kinetic parameters for the wave-form is not critical, as long as there is a certain time course. (3) The network is still grid-like as in previous detailed network, but the size of the network can be expanded in accordance with image resolutions, wherein each photoreceptor encodes single pixel. 

Thus, we constructed a photoreceptor based filter, the PR-filter, with only one tunable parameter—the gap junction strength. In addition, no inference or prior knowledge of noise is required. Therefore, it can be applied for blind denoising. 
Next, we tested the PR-filter on the Berkeley segmentation dataset(BSD68)\cite{martin2001database,zhang2017beyond} dataset and compared our results with classic spatial filters. In addition to a variety of regular noise types, we also generated two types of “blind-noise”. We listed both peak \emph{signal-to-noise ratio} (PSNR) and \emph{structural similarity index} (SSIM) of the denoised images. Surprisingly, we found the PR-filter dominants the SSIM performance on nearly all noise types (both high level and low level, Table 2,see Appendix 2 for parameter details), suggesting the PR-filter denoising aims to remain more details of the original images. The only exception to the PR-filter is the Salt\&Pepper noise. Even if to this particular noise type, the PR-filter achieves relative good performance in both PSNR and SSIM. As to
the PSNR index, the PR-filter achieves highest performance on nearly all types of noises in case the
noise levels are high, while it still remains good performance in case the noise levels are low (Table
1). In comparison, all other classic filters have weak performance when dealing with either regular
or blind noise types (Table 1). Overall, the PR-filter shows superior performance in SSIM index and excellent performance in PSNR index.

\subsection{PR-filter improves CNN blind denoising on non-Gaussian noise}

\begin{figure}[htbp]
  \centering
  \includegraphics[width=12cm]{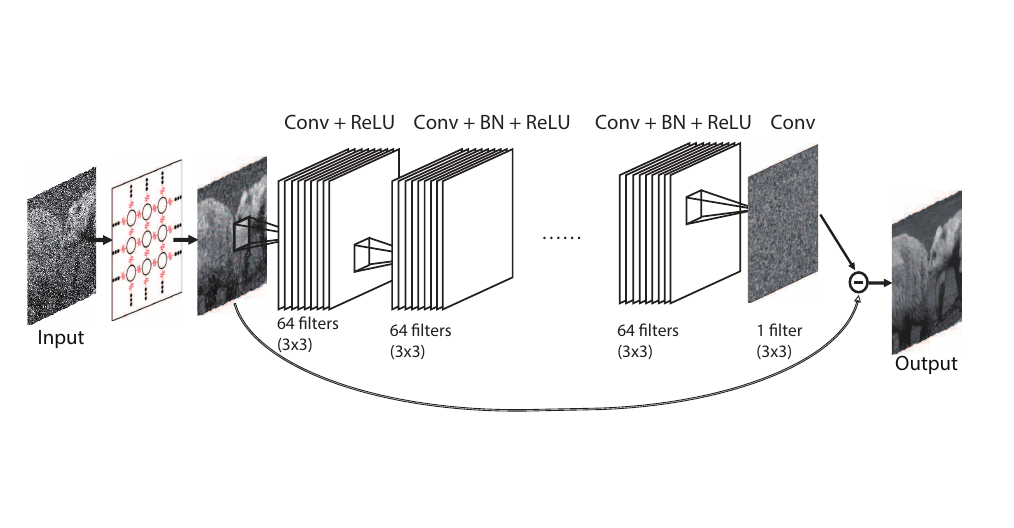}
  \caption{Architecture for PR filter + DnCNN. Noisy images are filtered by PR filter first before used as training data for DnCNN. We use 400 images of size 180$\times$180 for training\cite{chen2017trainable}. 128$\times$3,000 patches of size 50$\times$50 are cropped from the 400 images to train our model. For testing we use 68 natural images from BSD68. The weights are initialized by the method in \cite{abadi2016tensorflow}. We train our network using Adam optimizer with a initial learning rate of 0.001. Our network is trained for 50 epochs with a mini-batch size of 128. The learning rate is divided by 10 upon reaching 30 epochs. }
  \label{fig4}
\end{figure}

\begin{table}[htbp]
  \caption{blind denoising result of Network}
  \label{table3}
  \centering
  \begin{tabular}{ccccc}
    \toprule
    Type	&noisy	&DnCNN	&G-filter+CNN	&\emph{PR-filter+CNN}\\
    \midrule
\hline
\multirow{3}*{PSNR}	&9.5785 	&\color{red}24.1443	&22.7434	&\emph{23.5233}\\
	&12.4169 	&\color{red}26.1321	&25.1654	&\emph{25.5654}\\
	&15.2791 	&\color{red}27.8843	&27.1413	&\emph{27.4664}\\
\hline
\multirow{2}*{SSIM}	&0.0681 	&\color{red}0.7624	&0.7124	&\emph{0.7305}\\
	&0.1341 	&\color{red}0.8265	&0.7953	&\emph{0.8065}\\
	&0.2215 	&\color{red}0.8729	&0.8547	&\emph{0.8628}\\
    \bottomrule
  \end{tabular}
\end{table}

\begin{table}[htbp]
  \caption{blind non-Gaussian denoising result of Network}
  \label{table4}
  \centering
  \begin{tabular}{ccccc}
    \toprule
    Type	&noisy	&DnCNN	&G-filter+CNN	&\emph{PR-filter+CNN} \\
    \midrule
\hline
\multirow{3}*{PSNR}	&9.6814 	&21.93	&23.07	&\color{red}\emph{23.56}\\
	&14.1590 	&24.4234	&26.3956	&\color{red}\emph{26.5412}\\
	&17.7802 	&31.4521	&32.7721	&\color{red}\emph{33.5383}\\
\hline
\multirow{3}*{SSIM}	&0.0722 	&0.6884	&0.7197	&\color{red}\emph{0.7414}\\
	&0.1838 	&0.7762	&0.8307	&\color{red}\emph{0.8406}\\
	&0.3715 	&0.9357	&0.9526	&\color{red}\emph{0.9594}\\
    \bottomrule
  \end{tabular}
\end{table}

Most of the existing CNN blind denoising methods are essentially Gaussian denoising\cite{zhu2016noise}, which limits their general applications on non-Gaussian noise in reality. We wonder if combining the PR-filter with CNN denoising would improve the performance of the CNN blind denoising on non-Gaussian noise. We selected the framework “DnCNN”\cite{zhang2017beyond}, which is the state-of-the-art CNN blind denoising method, and re-implemented the DnCNN in Tensorflow\cite{abadi2016tensorflow}. The blind noises were identical to those in the section 3.2. The idea is to first process the raw noisy image by the PR-filter, then trained the DnCNN with the filtered image data (“PR-filter+DnCNN” , Fig.4). This architecture is similar to our visual system: noisy signals are first processed in the retina, and then sent to visual cortex for further improvements on details. 

Our results indicates the PR-filter+DnCNN outperforms original DnCNN by $\sim$2 dB in PSNR and 0.02-0.07 in SSIM while performing blind denoising on blind non-Gaussian noises (Table 4). Although the Gaussian filter can also improve the performance of DnCNN, it is less effective than the PR-filter (Table 4). Fig.5 illustrates the visual effects of the blind denoising on blind non-Gaussian noise. 
As to blind noises including Gaussian noise, DnCNN performs slightly better than PR-filter+DnCNN ($\sim$0.5 dB in PSNR and 0.01-0.02 in SSIM, Table 3), suggesting the PR-filter doesn’t significantly weaken the ability of CNN blind denoising on noises including Gaussian. 

Why the PR-filter can improve the performance of DnCNN on non-Gaussian blind noise? To gain insights into its working mechanism, we next analyzed the noise distributions before/after the PR-filter. Intriguingly, we found the PR-filter can regularize the noise distributions: In most cases we encountered, the filtered image noise was squeezed towards the center and formed distributions similar to Gaussian (Fig 6D-E, also see examples in Fig 6A and 6B). In very few cases where the original noise distributions appear discontinuous, the filtered image noise was still squeezed towards the center but formed distributions similar to mixture of Gaussian (Fig. 6C). The DnCNN adopted a strategy called “residual learning”, which aims to learn the distribution of the noise. As the DnCNN was initially optimized for Gaussian\cite{zhang2017beyond}, by changing irregular noise distributions closer to Gaussian, the PR-filter is able to improve the performance of DnCNN on non-Gaussian noise.
 
\begin{figure}[htbp]
  \centering
  \includegraphics[width=12cm]{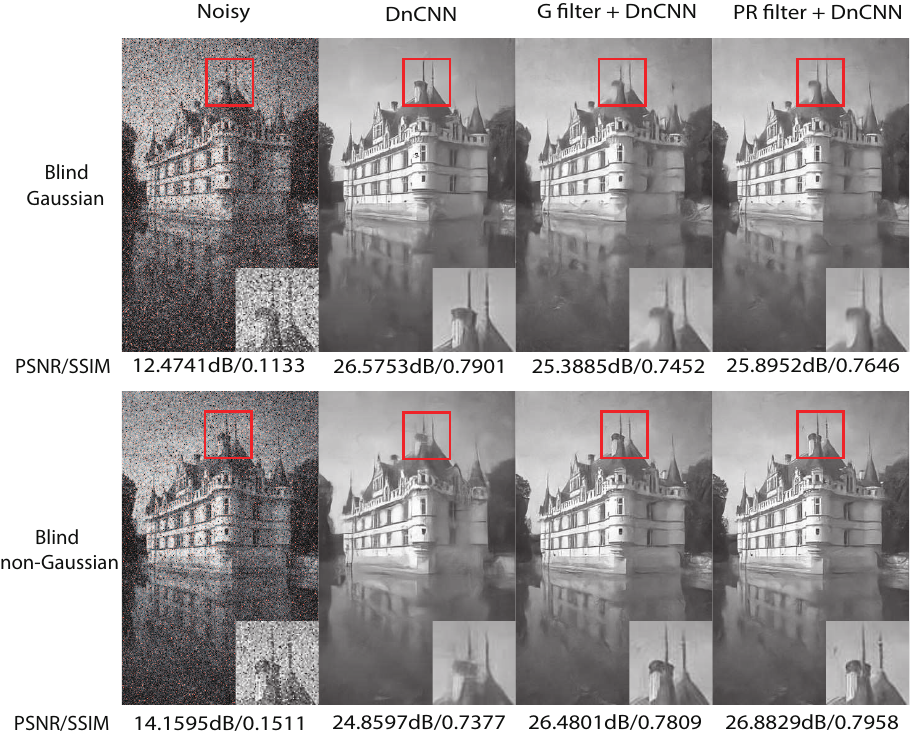}
  \caption{Example image for filter + DnCNN. Conductance for gap junctions in PR filter is 10nS and $\sigma$ for G filter is 2.}
  \label{fig5}
\end{figure}

\begin{figure}[htbp]
  \centering
  \includegraphics[width=13cm]{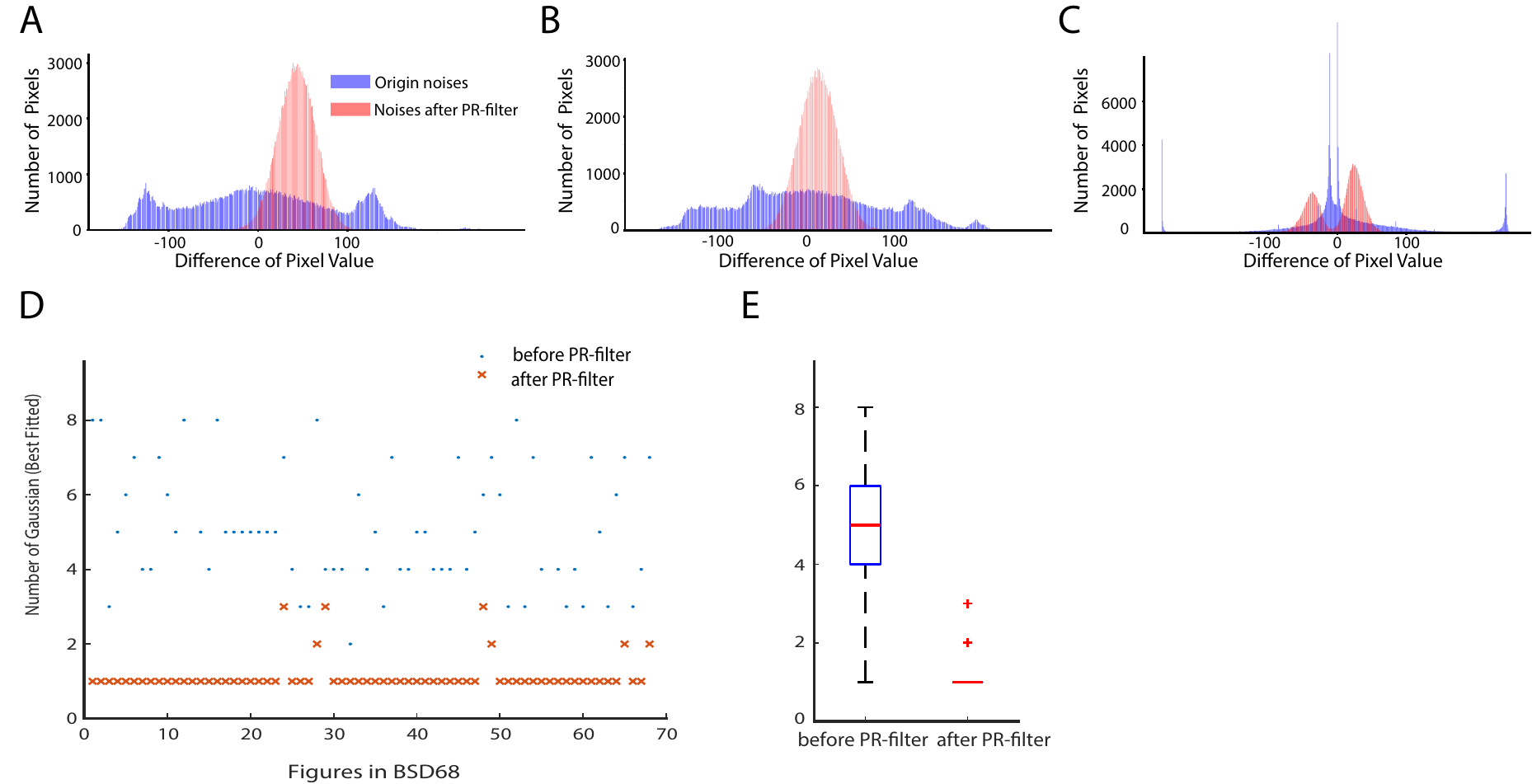}
  \caption{Noise distributions before and after the PR-filter. (A-C) Example noise distributions. (D) Fitting noise distributions before and after the PR-filter with Gaussian functions. (E) Statistics of fitted Gaussian numbers. Note most of the noise distribution after the PR-filter can be fitted by single Gaussian. }
  \label{a1supf4}
\end{figure}    

\section{Discussions}
In this study, inspired by the electrically coupled photoreceptors, we constructed a photoreceptor-filter (PR-filter). Although this simple filter contains only one variable, it exhibits excellent performance in blind denoising. In particular, it outperforms all classic spatial filters in SSIM index (with the only exception when dealing with Salt\&Pepper noise). In addition, it also significantly improves the performance (in both PSNR and SSIM) of the state-of-the-art CNN blind denoising on non-Gaussian noise. Why the PR-filter can keep good details of the image while efficiently removing noise? One plausible explanation is the receptive field revealed by STA analysis, which is similar to a narrow Gaussian filter. The photoreceptor in the PR-filter mostly sums up information from a very small patch, thus, it is more sensitive to details such as edges and lines. Why the PR-filter negatively impacts the performance of DnCNN on blind Gaussian noise? The reasons  might be: (1) we haven’t done extensive parameter search for the gap junction value. (2) the original noise distribution of blind Gaussian noise is already close to Gaussian. Can PR-filter also improve the performance of BM3D\cite{dabov2006image}? BM3D is essentially a non-blind denoiser and requires estimation of “noise level” ($\sigma$). Its performance might be more sensitive to the estimated noise level, rather than distribution of noise itself. As we expected, the PR-filter greatly improve the performance of BM3D given $\sigma$=25, but slightly negatively impact BM3D given $\sigma$=50 (Appendix Table A3.1).

\clearpage
\section*{Appendix1}
\setcounter{figure}{0} 
\renewcommand{\thefigure}{A1.\arabic{figure}}
\begin{figure}[htbp]
  \centering
  \includegraphics[width=13cm]{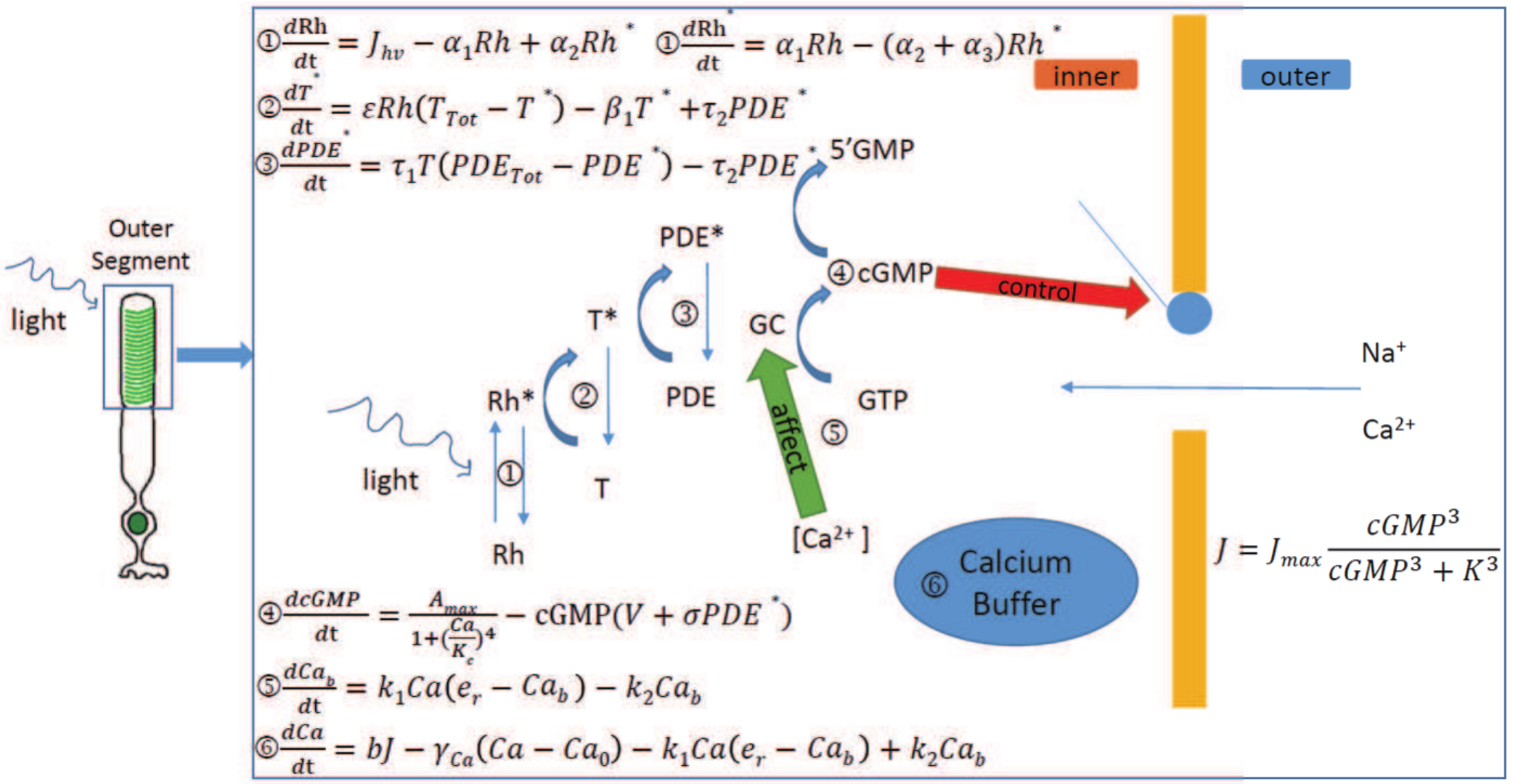}
  \caption{The mechanism of phototransduction}
  \label{a1supf1}
\end{figure}

The phototransduction cascade model was adapted from \cite{forti1989kinetics,torre1990model,kamiyama1996ionic} and we retuned the parameters in the NEURON simulator. The model framework is illustrated in Figure A1.1 and parameters are listed in Table A1.1, Table A1.2 and Table A1.3.

\begin{figure}[htbp]
  \centering
  \includegraphics[width=6cm]{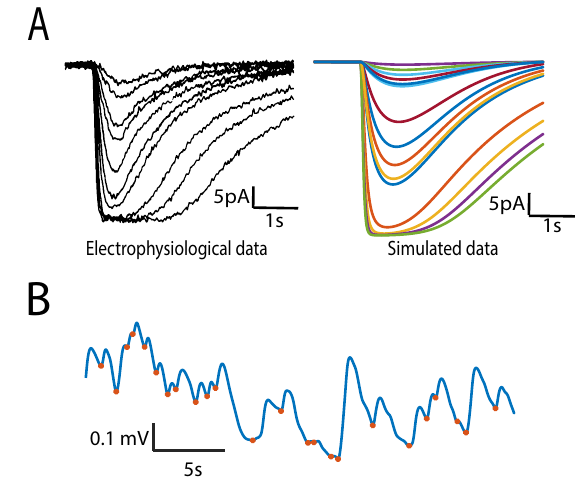}
  \caption{Photoreceptor network model for STA analysis. (A) Photocurrents in physiological data (left, adapted from \cite{baylor1986electrical}) and our detailed model (right). (B) Sample voltage response trace under white Gaussian noise. Red marker donates the peak point (viewed as ‘spike’) in trace.}
  \label{a1supf2}
\end{figure}

\setcounter{table}{0} 
\renewcommand{\thetable}{A1.\arabic{table}}

\begin{table}[htbp]
  \caption{The parameter of phototransduction mechanism}
  \label{a1supt1}
  \centering
  \begin{tabular}{cc}
    \toprule
    Parameter	&Value\\
    \midrule
$\alpha_{1}$	&0.05  $ms^{-1}$\\
$\alpha_{2}$	&0.0000003 $ms^{-1}$\\
$\alpha_{3}$	&0.00003 $ms^{-1}$\\
$\varepsilon$	&0.0005 $uM^{-1}ms^{-1}$\\
$T_{Tot}$	&1000 $uM$\\
$\beta$	&0.00025 $uM ms^{-1} pA^{-1}$\\
$\tau_{1}$	&0.0002 $uM^{-1} ms^{-1}$\\
$\tau_{2}$	&0.005 $ms^{-1}$\\
$PDE_{Tot}$	&100 $uM$\\
$\sigma$	&0.001 $uM^{-1} ms^{-1}$\\
$A_{max}$	&0.0656 $uM ms^{-1}$\\
$K_{c}$	&0.1  $uM$\\
V	&0.0004 $ms^{-1}$\\
$k_{1}$	&0.0002 $uM^{-1} ms^{-1}$\\
$k_{2}$	&0.0008  $ms^{-1}$\\
$e_{r}$	&500 $uM$\\
b	&0.00025 $uM ms^{-1} pA^{-1}$\\
$\gamma_{Ca}$	&0.05 $ms^{-1}$\\
$C_{a0}$	&0.1 $uM$\\
$J_{max}$	&5040 $pA$\\
K	&10 uM\\
    \bottomrule
  \end{tabular}
\end{table}

\begin{table}[htbp]
  \caption{The initial parameter of phototransduction differential equation}
  \label{a1supt2}
  \centering
  \begin{tabular}{cc}
    \toprule
    Initial parameter	&Value\\
    \midrule
$Ca_{0}$    	&0.3 $uM$\\
$Ca_{b0}$	&34.9 $uM$ \\
$cGMP^{*}_{0}$ 	&2$uM ms^{-1}$\\
    \bottomrule
  \end{tabular}
\end{table}

For modeling the whole rod photoreceptor, we add different ion mechanisms\cite{kamiyama1996ionic,barnes1989ionic,kourennyi2004reciprocal,liu2004effects,publio2006realistic} in rod soma, which are listed in Table A1.3.

\begin{table}[htbp]
  \caption{The ion mechanism of photoreceptor}
  \label{a1supt3}
  \centering
  \begin{tabular}{cc}
    \toprule
Mechanism &Conductance Value\\
\midrule
Calcium channel &2 $mS/cm^{2}$\\ 
Ca-dependent potassium channel &0.5 $mS/cm^{2}$\\
Delayed rectifying potassium channel &2.0 $mS/cm^{2}$\\
Noninactivating potassium channel  &0.85 $mS/cm^{2}$\\
Nonselective cation channel channel &3.5 $mS/cm^{2}$\\
Leak &0.6 $mS/cm^{2}$\\
Calcium Buffer &Not applicable\\
    \bottomrule
  \end{tabular}
\end{table}

\clearpage
\section*{Appendix2}
\setcounter{figure}{0} 
\renewcommand{\thefigure}{A2.\arabic{figure}}
\setcounter{table}{0} 
\renewcommand{\thetable}{A2.\arabic{table}}

For testing the performance of PR-filter, we used five common noises, including white Gaussian, intensity denpendent Gaussian, Laplacian, Salt\&Pepper, and uniform, and we mixed these noises pixelwisely and generated two kinds of blind noises. 

For blind noise including Gaussian noise, we generated blind noises by adding noises pixelwisely in the sequence of Gaussian, intensity denpendent Gaussian, Laplacian, Salt\&Pepper, and uniform, where we selected parameters randomly, but we made the average PSNR of noisy images to be $\sim$9, $\sim$12,and $\sim$15 respectively.

For blind noise excluding Gaussian noise, we made the average PSNR of noisy images to be $\sim$9, $\sim$14,and $\sim$18 respectively.

\section*{Appendix3}
\setcounter{figure}{0} 
\renewcommand{\thefigure}{A3.\arabic{figure}}
\setcounter{table}{0} 
\renewcommand{\thetable}{A3.\arabic{table}}

\begin{table}[htbp]
\tiny
  \caption{Denoising result of BM3D}
  \label{a1supt4}
  \centering
  \begin{tabular}{ccccccccccc}
    \toprule
    \multirow{3}{*}{Type}&\multirow{2}{*}{blind}&\multirow{2}{*}{blind}&\multicolumn{2}{|c|}{blind Gaussian} &\multicolumn{2}{c}{blind non-Gaussian} &\multicolumn{2}{|c|}{blind Gaussian} &\multicolumn{2}{c}{blind non-Gaussian}\\
    \multicolumn{1}{c}{}&\multirow{2}{*}{Gaussian}&\multirow{2}{*}{non-Gaussian}&\multicolumn{4}{|c|}{$\sigma_{BM3D}$=25}&\multicolumn{4}{|c}{$\sigma_{BM3D}$=50}\\
    \multicolumn{3}{c}{}&\multicolumn{1}{|c}{BM3D}&PR+BM3D&\multicolumn{1}{|c}{BM3D} &PR+BM3D &\multicolumn{1}{|c}{BM3D}&PR+BM3D&\multicolumn{1}{|c}{BM3D}&\multicolumn{1}{c}{PR+BM3D}\\
    \midrule
\hline
\multirow{3}*{PSNR}	&9.5785  &9.6814	&10.3207	    &19.7015    &10.4458	    &19.7273 &15.2713 &19.7419 &15.6448 &19.7933\\
	                &12.4169 &14.1590	&13.9610		&21.6104    &16.7682 	&23.2251 &22.7856 &22.7088 &24.4656 &24.1791\\
	                &15.2791 &17.7802	&19.1426		&23.9680    &22.9378		&25.3970 &24.7373 &24.5023 &25.5649 &25.3711\\
\hline
\multirow{3}*{SSIM}	&0.0681  &0.0722 	&0.0758	        &0.4623     &0.0805	    &0.4740 &0.1519 &0.4761 &0.1625 &0.4853\\
	                &0.1341  &0.1838	&0.1653	        &0.5400     &0.2618	    &0.5825 &0.5723 &0.5999 &0.6518 &0.6454\\
	                &0.2215  &0.3715	&0.3706		    &0.6274     &0.5811      &0.6957 &0.6561 &0.6650 &0.6702 &0.6752\\
    \bottomrule
  \end{tabular}
\end{table}

\bibliographystyle{unsrt}
\bibliography{ms}

\end{document}